\documentclass{article}

\PassOptionsToPackage{numbers, compress}{natbib}

\usepackage[preprint]{nips_2018}
\usepackage[hidelinks]{hyperref} 

\usepackage[utf8]{inputenc} 
\usepackage[T1]{fontenc}    
\usepackage{url}            
\usepackage{booktabs}       
\usepackage{amsfonts}       
\usepackage{nicefrac}       
\usepackage{microtype}      

\usepackage{amsmath}	    
\usepackage{graphicx}       
\usepackage{subcaption}     
\usepackage{enumitem,kantlipsum}
\usepackage{blindtext}
\usepackage{bm}
\usepackage[justification=centering]{caption}

\title{Modularity Matters: Learning Invariant Relational Reasoning Tasks}

\author{
  Jason Jo \\
  MILA, Universit\'{e} de Montr\'{e}al \\
  IVADO \\
  \texttt{jason.jo.research@gmail.com} \\
  \And
   Vikas Verma  \thanks{Work done while author was visiting Montreal Institute for Learning Algorithms.}\\
  Department of Computer Science\\
  Aalto University, Finland \\
  \texttt{vikas.verma@aalto.fi} \\   
  \And
   Yoshua Bengio \thanks{CIFAR Senior Fellow} \\
  MILA, Universit\'{e} de Montr\'{e}al \\
  CIFAR\\
}

\begin{document}

\maketitle

\begin{abstract}
In this article we focus on two supervised visual reasoning tasks whose labels encode a semantic relational rule between two or more 
objects in an image: the MNIST Parity task and the colorized Pentomino task. The objects in the images undergo random 
translation, scaling, rotation and coloring transformations. Thus these tasks involve \emph{invariant relational reasoning}. We observed 
uneven performance of various deep convolutional neural network (CNN) models on these two tasks. For the MNIST Parity task, we report that 
the VGG19 model soundly outperforms a family of ResNet models. Moreover, the family of ResNet models exhibits a general sensitivity to 
random initialization for the MNIST Parity task. For the colorized Pentomino task, now both the VGG19 and ResNet models exhibit 
sluggish optimization and very poor test generalization, hovering around 30\% test error. The CNN models we tested all learn hierarchies 
of fully distributed features and 
thus encode the \emph{distributed representation prior}. We are motivated by a hypothesis from cognitive neuroscience which posits that the 
human visual cortex is \emph{modularized} (as opposed to \emph{fully distributed}), and that this modularity allows the visual cortex to 
learn higher order invariances. To this end, we consider a modularized variant of the ResNet model, referred to as a \emph{Residual Mixture 
Network} (ResMixNet) 
which employs a mixture-of-experts architecture to interleave distributed representations with more specialized, modular representations. 
We show that very shallow ResMixNets are capable of learning each of the two tasks well, attaining less than 2\% and 1\% test error on 
the 
MNIST Parity and the colorized Pentomino tasks respectively. Most importantly, the ResMixNet models are extremely parameter efficient: 
generalizing better than various non-modular CNNs that have over 10x the number of parameters. These experimental results
support the hypothesis that modularity is a robust prior for learning invariant relational reasoning.
\end{abstract}

\section{Introduction}

The human visual system is able to learn discriminative representations for high level abstractions in the data that are also invariant to 
an incredibly large and varied collection of transformations \cite{GemanInvariance, PoggioInvariance}. A central question in computer 
vision is how to learn such representations. The current de-facto standard visual learning models are deep convolutional neural networks 
(CNNs) \cite{CNN, Neocognitron}. Deep CNNs have achieved an incredible amount of success in learning various visual tasks: object 
recognition \cite{AlexNet, VGG19, Inception1, ResNet}, segmentation \cite{RCNN,FullyConvSegmentation} and even visual question answering 
\cite{RelationalNetwork, NeuralModuleAndreas, HierarchicalAttention}. While various CNN models are able to exhibit record 
breaking, sometimes superhuman test generalization performance, it should be noted that this test generalization is in the 
\emph{identically 
and independently distributed (i.i.d) setting}. So-called adversarial noise has been shown to break various models on the aforementioned 
tasks \cite{AdvExamples, SegmentationAdvExamples, AdvSegmentation, FoolingVQA}, exposing the sensitivity of these models in what we refer to 
as the \emph{out-of-distribution (o.o.d) setting}. 
Therefore the search for simultaneously discriminative and highly invariant representations continues. 

While there are many qualitatively different deep CNNs in the literature, the majority of them can be interpreted as learning deep 
hierarchies of 
fully distributed features: for features $f_l^1, f_l^2$ at level $l$ of the hierarchy, these features $f_l^1, f_l^2$ get applied to the 
\emph{same} input $y_{l-1}$. Thus many deep CNN models encode the \emph{fully distributed representation prior.} In 
this article, we explore the efficacy of the fully distributed representation prior for learning invariant relational rules. To 
measure this, we focus on two relational reasoning tasks: a newly crafted MNIST Parity task and a colorized variant of the Pentomino task 
\cite{KnowledgeMatters}. These two tasks are supervised visual reasoning tasks whose labels encode a semantic (high-level) relational rule 
between two or 
more objects in an image. Most importantly, the objects in the image undergo a wide range of transformations: random translation, scaling, 
rotation and coloring. Therefore performance on these two tasks will measure a model's ability to learn an invariant relational reasoning 
rule between explicit objects. 

For these two tasks we found that conventional deep convolutional architectures did not perform well, which stimulated our quest for 
architectures incorporating a different kind of prior. We tested the VGG19 model \cite{VGG19} with batch-normalization \cite{BatchNorm} and 
ResNet models \cite{ResNet} 
with depth varying from 26 all the way to 152. For the  MNIST Parity task, we report that the VGG19 model soundly outperforms a family 
of ResNet models, training faster and generalizing better. In particular, the family of ResNet models exhibited a general sensitivity 
to choice of random seed for weight initialization. However, for the colorized Pentomino task, both the VGG19 and family of ResNet models 
perform poorly, exhibiting sluggish optimization and very poor generalization, with average test error hovering around 30\% across the 
tested models. 

To address the shortcoming of the tested deep CNNs on the colorized Pentomino task, we appeal to a hypothesis from cognitive neuroscience 
which posits that the human visual cortex is \emph{modularized}  \cite{CorticalAreaBody, 
Facevsobject} (as opposed to \emph{fully distributed}), and that this modularity allows the visual cortex to learn higher order 
invariances \cite{PoggioModularityInvariance, PoggioFaceModularity}. This paper is a first exploration towards a different style of 
architecture which would better reflect this prior. To this end, we consider a modularized variant of the ResNet model, referred to as 
\emph{Residual Mixture Networks} (ResMixNets) 
which employs a mixture-of-experts architecture  \cite{MixtureofExperts, GoogleMixturePaper} to interleave distributed representations with 
more specialized, modular representations. 

Our main empirical result is that we can deploy extremely parameter efficient ResMixNets that outperform  both the VGG19-BN and the family 
of ResNets 
for the MNIST Parity and colorized Pentomino tasks. The best non-modularized model we tested for the MNIST Parity task was the VGG19-BN 
architecture, which achieves an average test error of 2.27\% while having over 20 million parameters. Using a ResMixNet with only 274K 
params (over a 70x reduction in parameter count), we are able to achieve an average test of error 1.98\%. For the colorized Pentomino task, 
we were able to deploy a ResMixNet with 193K parameters that achieved 0.88\% average test error (almost a 30x reduction in test error). In 
light of the extreme parameter efficiency 
and stellar generalization performance of the ResMixNet on both the MNIST Parity and colorized Pentomino task, \emph{we conclude that these 
experimental results support the hypothesis that modularity is a robust prior for learning invariant relational reasoning.}

\section{Invariant Relational Learning Tasks}

Here we introduce the MNIST Parity task (Section~\ref{sec:mnist}) and a colorized variant of the Pentomino task originally introduced in 
\cite{KnowledgeMatters} (Section~\ref{sec:pentomino}). Both of these tasks are binary supervised tasks whose labels encode a semantic 
relational rule between two or more objects in the image. We view both of these tasks as requiring a machine learning model to learn higher 
order invariances because the objects in the images can undergo random translation, scaling, rotation and coloring transformations 
without changing the image label. In addition to the large number of invariances present in each of the datasets, we further challenge any 
proposed machine learning model by restricting the training set size. Thus we are operating in the high invariance, low sample regime.

\subsection{MNIST Parity Dataset} 
\label{sec:mnist}

\begin{figure}
    \centering
    \begin{subfigure}[b]{0.3\textwidth}
        \includegraphics[width=\textwidth]{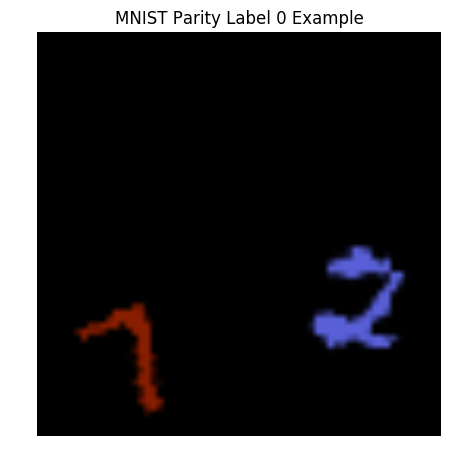}
    \end{subfigure}
    \begin{subfigure}[b]{0.3\textwidth}
        \includegraphics[width=\textwidth]{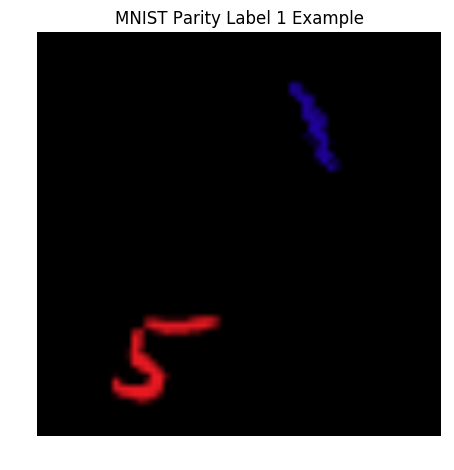}
    \end{subfigure}
    \caption{ \textbf{(Left)} Label 0 example: MNIST digit pair $\{2,7\}$ with different parity (one odd digit, one even digit) and 
\textbf{(Right)} Label 1 example: MNIST digit pair $\{1,5\}$ of the same parity (both odd digits). Digits are subject to random 
translations, scalings, rotations and coloring. Best viewed in color. }\label{fig:mnist-parity-examples}
\end{figure}

The MNIST Parity dataset consists of 30K training, 5K validation and 5K test images. Each image is of size 64$\times$64 and is divided into 
a 2$\times$2 grid 
of 32$\times$32 blocks. Each image has two 28$\times$28 MNIST digits placed in 2 randomly chosen blocks out of these four blocks. A digit is 
randomly 
colored (using one out of 10 randomly chosen colors), randomly scaled to size $\in \{20\times 20, 22 \times 22, 24 \times 24, 26 \times 26, 
28 \times 28\}$, randomly 
rotated by angle $\theta \in \{0, 5, 10, 15, 20, 25, 30\}$ and placed at a random location with in a block. The task is to predict whether 
both the digits in an image are of the same parity, both even or both odd (label 1) or not (label 0). Example images are shown in 
Fig.~\ref{fig:mnist-parity-examples}. 
We note that the MNIST Parity training, validation and test images are generated and deformed from the original MNIST training, validation 
and test digits respectively.

\subsection{Colorized Pentomino Dataset}
\label{sec:pentomino}

\begin{figure}
    \centering
    \begin{subfigure}[b]{0.3\textwidth}
        \includegraphics[width=\textwidth]{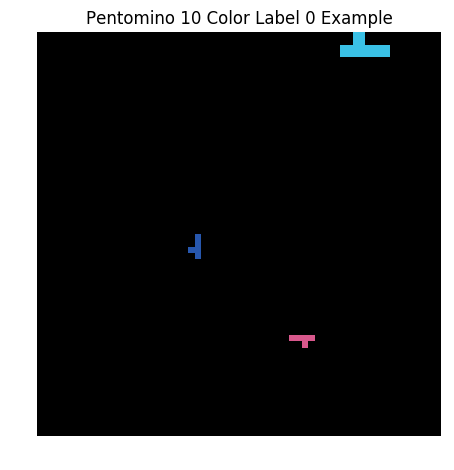}
    \end{subfigure}
    \begin{subfigure}[b]{0.3\textwidth}
        \includegraphics[width=\textwidth]{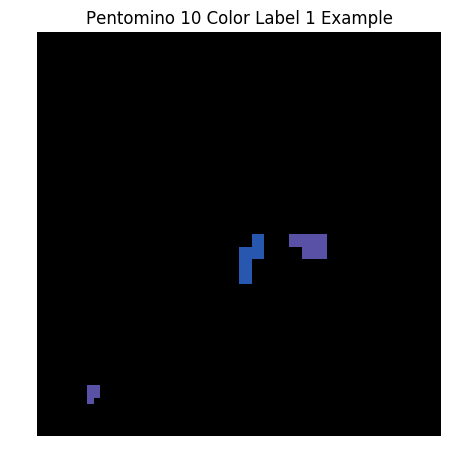}
    \end{subfigure}
    \caption{ \textbf{(Left)} Label 0 example (all the shapes are of the same sprite type) and \textbf{(Right)} Label 1 example (there 
exists a sprite of a 
    \emph{different type} than the other sprites). Sprites are subject to random translations, scalings, rotations and coloring. Best viewed 
in color. }\label{fig:pentomino-10-color-examples}
\end{figure}

The Colorized Pentomino dataset consists of 20K train, 5K validation and 5K test images. Each image is of size 64$\times$64, which is 
divided 
into a grid of 8$\times$8 blocks. Each image has 3 Pentomino sprites placed in 3 randomly chosen unique blocks. The Pentomino sprite type, 
scaling amount and rotation angles are the same as in \cite{KnowledgeMatters}. Additionally, we color the Pentomino sprites randomly using 
one out of 10 colors. Due to the extra coloring transformation, the colorized Pentomino has 10x the number of invariances as the original 
Pentomino dataset. As in \cite{KnowledgeMatters}, the task is to learn whether all the Pentomino sprites in an image belong to same class 
(label 0) or not (label 1). Example images are shown in Fig.~\ref{fig:pentomino-10-color-examples}.

Relational object reasoning tasks have two key defining characteristics: the object distribution and the relational rule. In this manner, 
we believe that the MNIST Parity task and the colorized Pentomino task are qualitatively different. The MNIST Parity task consists of 
curvilinear digit strokes while the colorized Pentomino task consists of rigid polygonal shapes. With respect to the relational rule, the 
MNIST Parity task is an AND operation on the parity of the digits while the colorized Pentomino task is a XOR like operation on the 
sprite types. The two datasets furthermore differ in following aspects: Colorized Pentomino has more sparsity in the images and the 
objects in the image have more freedom for translation as compared to MNIST Parity. Furthermore, all the objects in the Colorized Pentomino 
dataset are made of only straight edges, whereas the MNIST Parity dataset consists of different types of curves. Arguably, these curves 
assist 
more than the straight edges of Colorized Pentomino dataset in learning discriminative features for the desired task.

\subsection{Psychological Tests of Relational Reasoning}
While both the datasets are artificial, we note that in the field of psychology and human intelligence, similar visual tasks have been used 
to measure relational reasoning in human beings \cite{RelationalReasoningPsych, CultureFreeIntelligenceTest, QuoteTheRaven}. Our tasks are 
also completely ``figural'' in that no outside information is needed to be able to solve these problems. Indeed these tasks are designed to 
measure what \cite{QuoteTheRaven} refers to as \textbf{eductive ability}: 
\begin{quote}``\emph{...the ability to make meaning out of confusion, the ability to generate high-level, usually nonverbal, schemata which 
make it easy to handle complexity}.'' 
\end{quote}

\section{Modularity and Relational Reasoning}

\subsection{Why modularize?}

In this article we are interested in invariant relational learning. In this setting, a machine learning model must be able to recognize 
that simply translating, rotating, scaling or changing the color of any of the objects in the image does not change the label of the image. 
Therefore, any proposed machine learning model will be tasked with learning simultaneously discriminative and invariant representations 
\cite{GemanInvariance, PoggioInvariance}. We are motivated by the following question: \emph{which architectural priors will facilitate 
learning of such representations?} 

The de-facto visual learning models used today are deep CNNs. Many of these deep CNNs may be classified as learning a 
deep hierarchy of fully distributed features: for features $f_l^1, f_l^2$ at level $l$ of the hierarchy, these features $f_l^1, f_l^2$ get 
applied to the \emph{same} input $y_{l-1}$. Overall, distributed representations \cite{HintonDistReps} have been an extremely 
powerful architectural prior for AI. However, when the number of invariances in the dataset is very large (and/or the dataset size is 
sufficiently small), one may encounter the 
\emph{interference problem} \cite{BeckerHinton, InterferenceBook, MixtureofExperts,ToModularize} for architectures that learn fully 
distributed representations. In the 
case of supervised learning from image labels, there is one global teaching signal, and \cite{BeckerHinton} conjectured that this would 
entangle all the neural network's parameters, which would cause the features to interfere with one another and result in a slow down in 
learning. Similarly in \cite{InterferenceBook}, it is hypothesized that a neural network with a ``\emph{homogeneous connectivity}'' 
topology 
(e.g. encoding the fully distributed representation prior) will struggle to simultaneously learn many different patterns, as each pattern 
will interfere with each other. When a dataset has a large number of invariances, a machine learning model must learn to associate a large 
number of seemingly unrelated patterns with one another, which may exacerbate the interference problem. Take for example 
the MNIST Parity task: a machine learning model must learn associate the digit pairing $\{1, 4\}$ with $\{2, 7\}$ as they have the same 
label of 0, but the digit pairings have different geometric properties. 

One natural way to combat the interference problem is to allow for specialized sub-modules in our architecture. Once we 
modularize, we reduce the amount of interference that can occur between features in our model. These specialized modules can now learn 
highly discriminative yet invariant representations while not interfering with each other \cite{PoggioModularityInvariance, 
PoggioFaceModularity}. To this end, there is much supporting evidence from the cognitive neuroscience research: the Fusiform Face Area 
(FFA) \cite{FusiFormFaceAreaOriginalPaper}, face selective cells \cite{FaceSelectiveCells}, and cells that are selective only to portions of 
the 
human body as opposed to the human face \cite{CorticalAreaBody}. \emph{In the case of invariant relational learning, we hypothesize that 
modularity allows for the development of specialized neural circuitry that can learn to associate many seemingly unrelated patterns.}

\subsection{Residual Mixture Network} 

A well established neural architecture to combat the interference problem is the so-called \emph{mixture of experts} (MoE)
\cite{MixtureofExperts}. There are two key components to the MoE architecture:
\begin{enumerate}
 \item Individual expert networks $\{E_1, \dots, E_n\}$ (which here map their input to their output). 
 \item A Gater network $G$ that weights the output from each of the individual experts, in a way that is context-dependent. 
\end{enumerate}

\begin{figure}
    \centering
    \begin{subfigure}[b]{0.7\textwidth}
         \centering
         \includegraphics[scale=0.5]{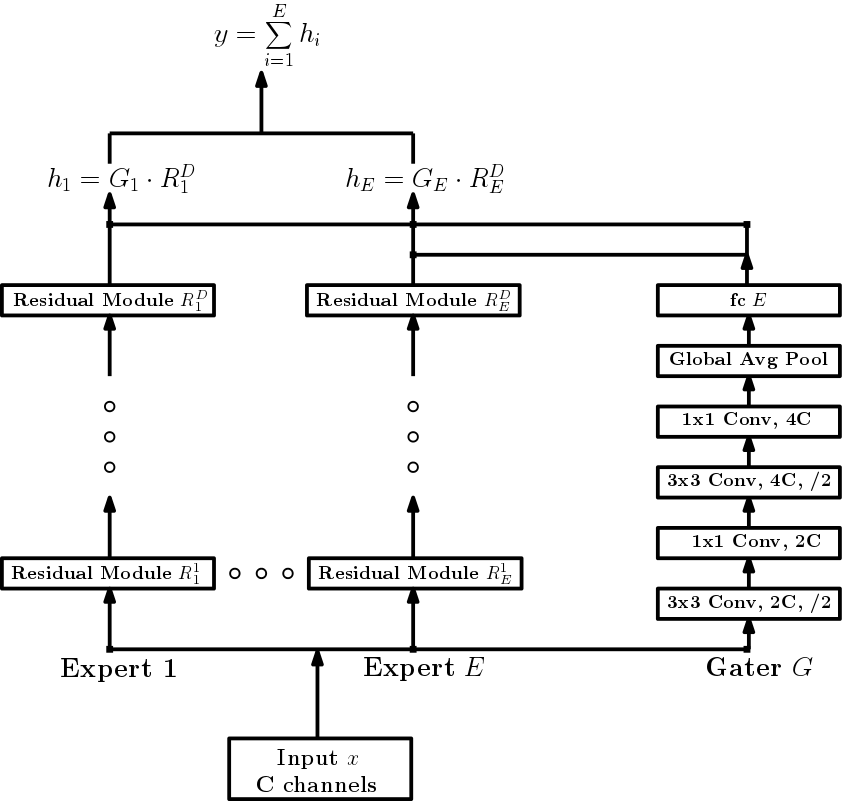} \caption{$M(E,D)$: A mixture of $E$ experts, where each 
expert is a $D$ stack of residual modules and a gater network $G$ which weights all the experts and forms an additive 
mixture.}\label{fig:resmixnet-module}
    \end{subfigure} \hfill
    \begin{subfigure}[b]{0.28\textwidth}
    \centering
        \includegraphics[scale=0.3]{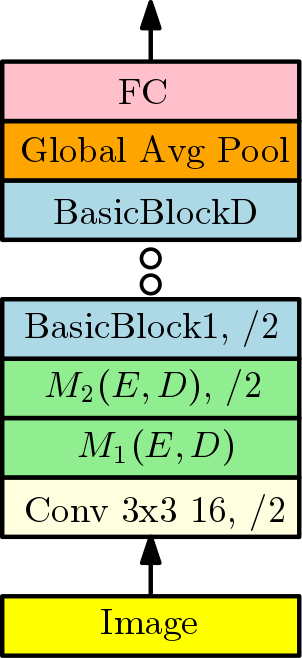}\caption{ResMixNet($E,D$) 
consisting of $D$ stacks of residual modules and $E$ experts. Best viewed in color.}\label{fig:resmixnet-diagram}
    \end{subfigure}
\end{figure}

In Fig.~\ref{fig:resmixnet-module} we present our mixture of experts module. Each of our experts $E_i$ is a $D$ stack of 
residual modules, specifically the Basic Block from \cite{ResNet}. Our Gater network $G$ is a stack of four convolutional layers, a global 
avg pooling layer and a dense layer with a softmax activation. Thus the output of the gater network $G$ is an $E$-length probability vector 
and we use this to output a weighted linear combination of the experts, e.g. the output $y = \sum_{i=1}^E G[i] E_i$.

We present our Residual Mixture Network (ResMixNet) architecture in Fig.~\ref{fig:resmixnet-diagram}. We observe that 
the ResMixNet interleaves distributed and modular representations together. For example, the first layer in our network is a fully 
distributed convolutional layer with 16 3x3 filters with stride 2. We feel this is the appropriate prior for the MNIST Parity and colorized 
Pentomino tasks because each of 
those datasets share low level features, e.g. curvilinear digit strokes for MNIST Parity and straight edges for the colorized Pentomino. It 
would be wasteful to have each individual expert learn its own low level edge filters. Also note that each expert in our network receives 
the same input as any other expert, but then each expert learns its own specialized representation through its $D$ stack of residual 
modules. These modularized representations are then weighted together using the gater network $G$. We stack two expert modules $M_1$ and 
$M_2$ together and then have the third block of our network to be a $D$ stack of BasicBlock residual modules. Following the ResNet recipe, 
we have our second and third blocks reduce the 
spatial height and width dimensions by 2 by using a stride 2 convolution, hence the ``/2'' notation in 
Figs~\ref{fig:resmixnet-module}--\ref{fig:resmixnet-diagram}. Again following 
ResNet recipe, whenever the spatial dimension is reduced, we double the number of filters. Our final output layer is a 2 unit dense layer. 
We use a two 
class negative log likelihood loss function. 

\section{Experimental Results}

In all of our experiments we used a training batch size of 64 and L2 weight decay parameter of 1e-4. We train all of our models using 
SGD+Momentum 
with momentum parameter $\mu = 0.9$, we try 5 different learning rates $l \in \{0.1, 0.05, 0.01, 0.005, 0.001\}$ and perform 5 trials with 
seed $\in \{0,1,2,3,4\}$. We train for 200 epochs and decay the learning rate by a factor of 10 at epochs 100 and 140. We report the 
learning rate with the best average test performance. Due to the sparsity in the datasets, all of our networks have an initial 
convolutional layer with stride 2. We used two-class softmax and negative log likelihood as our loss function. We refer to the supplementary 
materials for further details. 

\subsection{MNIST Parity}

\begin{table}
  \caption{MNIST Parity Generalization Results}
  \label{table:mnist-parity-gen-table}
  \centering
  \begin{tabular}{lll}
    \toprule
    \textbf{Model}     		& \textbf{Parameter Count}     	& \textbf{Test Error} \\
    \midrule
    ResNet26 			& 370K		  		& $30.96 \pm 23.71\%$     \\
    ResNet50    		& 758K		 		& $30.36 \pm 23.34\%$      \\
    ResNet152-Bottleneck     	& 3.66M		       		& $11.24 \pm 19.03\%$  \\
    VGG19-BN     		& 20M		       		& $2.27 \pm 0.10\%$  \\
    ResMixNet(2,2)    		& \textbf{274K}		       	& \bm{$1.98 \pm 0.07\%$}  \\
    \bottomrule
  \end{tabular}
\end{table}

\begin{figure}
    \centering
    \includegraphics[width=0.5\textwidth]{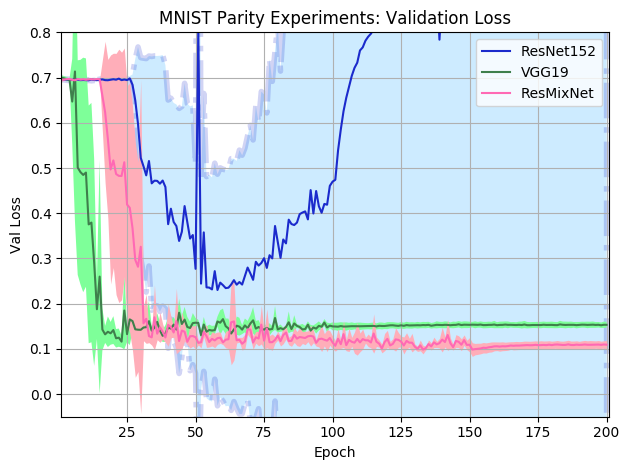}
     \caption{ Average Validation Loss of the best performing models: ResNet152, VGG19-BN and ResMixNet(2,2). 
}\label{fig:mnist-parity-val-loss-plot}
\end{figure}

In Table \ref{table:mnist-parity-gen-table} we present the generalization performance of the various models for the MNIST Parity task. We 
first note the somewhat surprising result that the VGG19-BN network soundly outperforms the ResNet models. To the best of our knowledge, 
this 
is the first time such a performance gap has been exhibited between a residual network and non-residual network. We witnessed a sensitivity 
to randomized initialization for all the ResNet models evidenced by the large standard deviation of the average test error. While the 
VGG19-BN 
does exhibit stellar performance, we note that our ResMixNet(2,2) model actually attains {\bf slightly better test performance while having 
over 
70x fewer parameters}. We see in Fig.~\ref{fig:mnist-parity-val-loss-plot} that the ResMixNet(2,2) model is able to obtain lower 
validation loss than the VGG19-BN model. 

\subsection{Colorized Pentomino}

\begin{table}
  \caption{Pentomino 10 Color Generalization Results}
  \label{table:pentomino-10-color-gen-table}
  \centering
  \begin{tabular}{lll}
    \toprule
    \textbf{Model}     		& \textbf{Parameter Count}     	& \textbf{Test Error} \\
    \midrule
    ResNet26 			& 370K		  		& $34.61 \pm 4.67\%$     \\
    ResNet50    		& 758K		 		& $30.31 \pm 1.74\%$      \\
    ResNet152-Bottleneck     	& 3.66M		       		& $31.02 \pm 3.06\%$  \\
    VGG19-BN     		& 20M		       		& $34.01 \pm 4.72\%$  \\
    ResMixNet(4,1)    		& \textbf{193K}		       	& \bm{$0.88 \pm 0.12\%$}  \\
    \bottomrule
  \end{tabular}
\end{table}

\begin{figure}
    \centering
    \begin{subfigure}[b]{0.45\textwidth}
        \includegraphics[width=\textwidth]{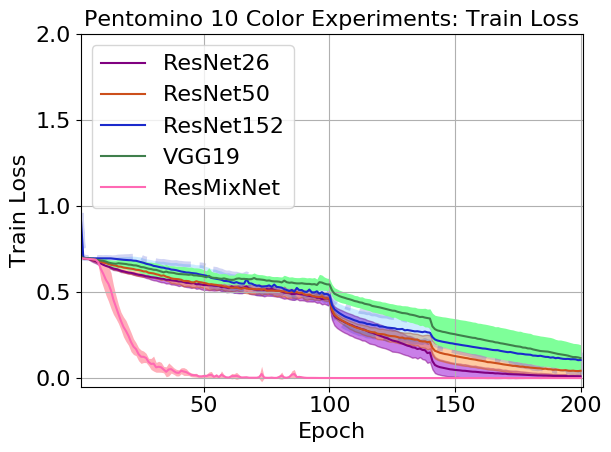}
		\label{fig:pentomino-10-color-train}
    \end{subfigure}
    \begin{subfigure}[b]{0.45\textwidth}
        \includegraphics[width=\textwidth]{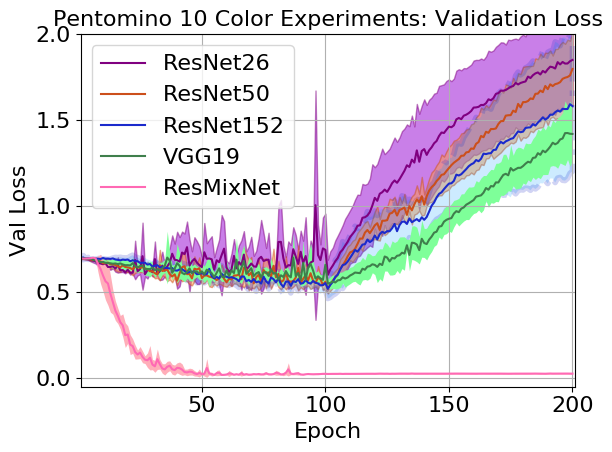}
  		\label{fig:pentomino-10-color-val}
    \end{subfigure}
    \caption{ \textbf{(Left)} Train Loss and \textbf{(Right)} Validation Loss 
performance on the Pentomino 10 color dataset. Best viewed in color. }\label{fig:pentomino-10-color}
\end{figure}

In Table~\ref{table:pentomino-10-color-gen-table} we present the generalization performance of the various models for the Pentomino 10 
color task. In contrast to the MNIST Parity results, we observe that now the VGG19-BN and the various ResNet models generalize poorly. 
From Fig.~\ref{fig:pentomino-10-color} we conclude that on this task the non-modularized networks have a tendency to 
slowly converge to poorly generalizing local minima. 

On the other hand, from the Table~\ref{table:pentomino-10-color-gen-table} and Fig.~\ref{fig:pentomino-10-color} we observe stellar 
optimization and generalization performance of the ResMixNet(4,1) model. This model is again able to outperform networks that have many 
more 
parameters. We highlight a {\bf nearly 30x reduction in test error} from the non-modularized CNNs to the ResMixNet(4,1) model.

\subsection{Classical Object Recognition}
Finally we present the performance of the ResMixtureNet for three object recognition tasks: CIFAR-10, CIFAR-100 and SVHN. We train as before 
except that now we decay the learning rate by a factor of 10 at 
epoch 100 and 150. For the SVHN we use a similar setup but with learning rate 0.01 and we train for only 40 epochs, reducing the learning 
rate by a factor of 10 after 20 and 30 epochs. For the CIFAR-10/100 datasets we augment with on-the-fly random horizontal flips and random 
cropping with pad size 4. We use random seed 0 for the initialization. The results of our experiments are shown in 
Table~\ref{table:object-recognition-gen-table}.

We set our baseline of performance as the ResNet50 and we test ResMixNet(5,3) which has roughly the same number of parameters. We notice 
from the table that the performance on CIFAR-10 is quite close, merely a $0.74\%$ gap in test error and that for SVHN that the performance 
of the two models is even closer, a mere difference of $0.13\%$. However we see that the gap is $5.46\%$ for 
the CIFAR-100. Note that for the CIFAR-100, the data by design has multiple class labels that are semantically similar, and thus many of the 
images may share features. In this case the ResMixNet may not be an optimal prior by itself. This suggests future work exploring 
combinations of this
prior with the classical CNN prior based on a sequence of transformations.

\begin{table}
  \caption{ CIFAR-10, CIFAR-100 and SVHN Results}
  \label{table:object-recognition-gen-table}
  \centering
  \begin{tabular}{lccc}
    \toprule
    \textbf{Model}   & \textbf{Dataset} &  \textbf{Num. Params}   & \textbf{Test Accuracy} \\ 
    \midrule
    ResNet50   & CIFAR-10  & 758K	& $93.48\%$    \\
    ResMixNet(5,3) & CIFAR-10 &  748K	& $92.74\% $  \\
    \midrule
    ResNet50 	& CIFAR-100 & 764K & $71.81\%$ \\
    ResMixNet(5,3) & CIFAR-100 &  754K	& $66.35\% $  \\
    \midrule
    ResNet50   & SVHN  & 758K	& $95.45\%$    \\
    ResMixNet(5,3) & SVHN &  748K	& $95.58\% $  \\
    \bottomrule
  \end{tabular}
\end{table}

\section{Related Work}

With respect to relational reasoning, there has been much recent work on Visual Question and Answering (VQA) \cite{VQA, 
CLEVR,VisualTuringTest, MultiWorld, VizWiz}. The MNIST Parity task and the Pentomino tasks have no question and answering component and are 
of a purely visual nature. The VQA tasks require learning multiple types of relations while each of the MNIST Parity and Pentomino tasks 
require learning a single invariant relational rule that applies to a set of objects and their many transformations. 

With respect to the ResMixNet architecture, it is clearly built on top of the well established mixture of experts architecture 
\cite{MixtureofExperts,BengioModularity}. The most relevant work is \cite{GoogleMixturePaper}. One major difference with our work is that 
\cite{GoogleMixturePaper} applied modular MLPs to a task like Jittered MNIST, which does not have as many invariances as the MNIST Parity 
and the colorized Pentomino task, nor does it involve any relational reasoning. To this end, we believe that the higher order invariance 
setting is where the modularity prior can result in a significant performance gain which we showed in Section 4. 

In general there has been some recent work on modular neural networks. In \cite{SparseMOE} a huge modular neural network is used to achieve 
state of the art performance on language modeling and machine translation tasks. The ResNeXt model \cite{ResNext} uses multi-branches (e.g. 
experts) and pools the experts together via summation, but they do not employ a gater-type network to weight the sum. The Inception 
architectures \cite{Inception1, Inception2, Inception4} also uses multi-branch modules and concatenates all them together, thus they 
similarly lack a gater network.

\section{Conclusion}
In this article we tested the performance of four well known types of CNN models (VGG19-BN, ResNet26, ResNet50 and ResNet152-Bottleneck) on 
two invariant relational reasoning tasks: the MNIST Parity task and a colorized variant of the Pentomino task. For these two tasks we 
observed that conventional deep CNN models did not perform well overall. We hypothesized that the root of the problem is the so-called 
\emph{interference problem}. For invariant relational reasoning tasks, a machine learning model must learn to associate a large number of 
seemingly unrelated patterns in the data. The interference problem posits that for models that employ fully distributed feature hierarchies, 
these patterns can interfere with one another and result in poorly conditioned training and/or suboptimal generalization.  

One natural remedy to the interference problem is to deploy machine learning models that learn modularized representations. To this end, we 
proposed a modularized CNN: the Residual Mixture Network (ResMixNet) which combines the existing mixture of experts architecture with the 
ResNet architecture. We showed that the ResMixNet is able to learn both the MNIST Parity and colorized Pentomino tasks, exhibiting less than 
2\% and 1\% test error, respectively. Most importantly the ResMixNets are able to outperform networks that have well over 10x the number of 
parameters. We believe our empirical results support the hypothesis that modularity can be a robust prior for learning invariant relational 
reasoning rules. 

Finally we tested the ResMixNets on three object recognition tasks: CIFAR-10, CIFAR-100 and SVHN. We used a ResNet50 model as our baseline 
of performance. We constructed a ResMixNet that has roughly the same parameter budget as the ResNet50. For the CIFAR-10 and SVHN 
classification tasks, the ResMixNet exhibited a less than 1\% gap in test error from the ResNet50 baseline. However, for the CIFAR-100, the 
ResMixNet lagged behind the ResNet50 model by over 5\%. We hypothesized that this is due to the fact that the CIFAR-100 dataset has many 
similar class labels and thus many of the images may share features. In this case, modularity would be a sub-optimal prior. 

For future work we aim to understand the optimal balance between modular and distributed representations, all towards the end goal of a 
robust general visual architecture that can learn from a wide array of data distributions (artificial, natural, relational reasoning 
oriented, object recognition oriented, etc.). 

\subsubsection*{Acknowledgments}
Vikas Verma was supported by Academy of Finland project 13312683 / Raiko Tapani AT kulut.
We would like to acknowledge the following organizations for their generous research funding and/or computational support (in alphabetical 
order): Calcul Qu\'{e}bec, Canada Research Chairs, the CIFAR, Compute Canada, the IVADO and the NSERC.

\appendix

\section{Optimization Experiments}

In this section we expand upon our experimental setup and share some more details. In the main article we listed our SGD+Momentum 
hyperparameter setup. In preliminary experiments we explored other optimizers over a range of settings. Specifically, we tested the 
Adadelta 
\cite{Adadelta} optimizer over learning rates $l \in \{0.1, 0.2,\dots, 1.0\}$ with $\rho=0.9$ and we also tried Adam \cite{Adam} with 
learning rates $l \in \{0.1, 0.01, 0.001, 0.0001\}$ with $\beta_1=0.9, \beta_2=0.999$ and we did not notice any non-trivial difference in 
performance. 

In Table~\ref{table:stride-2-lr-table} we list the LR for SGD+Momentum with the best average test accuracy over 5 random trials for the 
Colorized Pentomino and MNIST Parity tasks.

\begin{table}[h]
  \caption{Best LRs for SGD+Momentum}
  \label{table:stride-2-lr-table}
  \centering
  \begin{tabular}{lccc}
    \toprule
    \textbf{Model}   & \textbf{Dataset} & \textbf{LR}\\ 
    \midrule
    ResNet26 & Colorized Pentomino & 0.1 \\
    ResNet50 & Colorized Pentomino & 0.1 \\
    ResNet152 & Colorized Pentomino & 0.1 \\ 
    VGG19-BN & Colorized Pentomino & 0.1 \\ 
    ResMixNet(4,1) & Colorized Pentomino & 0.01 \\ 
    ResNet26 & MNIST Parity & 0.05 \\
    ResNet50 & MNIST Parity & 0.1 \\
    ResNet152 & MNIST Parity & 0.05 \\ 
    VGG19-BN & MNIST Parity & 0.01\\ 
    ResMixNet(2,2) & MNIST Parity & 0.1 \\ 
    \bottomrule
  \end{tabular}
\end{table}

We also ran experiments for the ResNet26, ResNet50, ResNet152-Bottleneck and VGG19-BN with stride 1 in the convolutional layer as opposed 
to 
stride 2 and there was no non-trivial difference in average test error. None of the models were able to display qualitatively different 
performance with respect to closing the gap of the ResMixNet's performance. 

\section{Choosing the Residual Mixture Network's hyperparameters}

The ResMixNet has two hyperparameters: the number of experts $E$, and since each expert $E$ is parameterized to be a deep stack of 
BasicBlock residual modules, the other hyperparameter is the depth $D$. Our only guiding principle was to never have the parameter count of 
our ResMixNet exceed the parameter count of the shallowest non-modular network we considered. Thus we enforced the constraint that we never 
exceed the parameter count of the ResNet26, which has 370K. This really limits the possible figurations of the ResMixNet.

\newpage

\small
\bibliographystyle{unsrt}


\end{document}